\documentclass[sigconf,nonacm]{acmart}
\usepackage{booktabs}
\usepackage{multirow}

\usepackage{listings}

\AtBeginDocument{%
  \providecommand\BibTeX{{%
    \normalfont B\kern-0.5em{\scshape i\kern-0.25em b}\kern-0.8em\TeX}}}

\begin{document}

\title{Pathway: a fast and flexible unified stream data processing framework for analytical and Machine Learning applications}

\author{Micha\l{} Bartoszkiewicz}
\affiliation{\country{}}

\author{Jan Chorowski}
\authornote{Authors in alphabetical order. Corresponding author: {jan.chorowski@pathway.com}.}
\affiliation{\country{}}

\author{Adrian Kosowski}
\affiliation{\country{}}

\author{Jakub Kowalski}
\affiliation{\country{}}

\author{Sergey Kulik}
\affiliation{\country{}}

\author{Mateusz Lewandowski}
\affiliation{\country{}}

\author{Krzysztof Nowicki}
\affiliation{\country{}}

\author{Kamil Piechowiak}
\affiliation{\country{}}

\author{Olivier Ruas}
\affiliation{\country{}}

\author{Zuzanna Stamirowska}
\affiliation{\country{}}

\author{\hspace*{0em}}
\email{{firstname.lastname}@pathway.com}

\affiliation{%
  \institution{Pathway.com}
  \city{Paris}
  \country{France}\vspace*{5mm}
}

\author{Przemys\l{}aw   Uzna\'nski}
\affiliation{\country{}}

\renewcommand{\shortauthors}{Pathway team} %

\begin{abstract}
  We present Pathway, a new unified data processing framework that can run workloads on both bounded and unbounded data streams. The framework was created with the original motivation of resolving challenges faced when analyzing and processing data from the physical economy, including streams of data generated by IoT and enterprise systems. These required rapid reaction while calling for the application of advanced computation paradigms (machine-learning-powered analytics, contextual analysis, and other elements of complex event processing). Pathway is equipped with a Table API tailored for Python and Python/SQL workflows, and is powered by a distributed incremental dataflow in Rust. We describe the system and present benchmarking results which demonstrate its capabilities in both batch and streaming contexts, where it is able to surpass state-of-the-art industry frameworks in both scenarios. We also discuss streaming use cases handled by Pathway which cannot be easily resolved with state-of-the-art industry frameworks, such as streaming iterative graph algorithms (PageRank, etc.). 
\end{abstract}



\maketitle

\section{Introduction}
Traditionally, data processing systems were designed either for high throughput batch computations, or for low latency streaming processing. However, modern data applications often demand low latencies at high data throughputs. One solution is the lambda architecture  \cite{lambda}, which calls for running two similar workloads: a batch one for exact computations on historical data and a streaming one used to patch the batch results with latest data. Alternatively, aiming to avoid architecture complexity, it is also possible to rely on data processing frameworks which unify batch and streaming computations. 

The new data processing framework which we describe in this paper, Pathway, has a unified runtime suitable for running both streaming and batch tasks. Its design results from the need to perform certain types of real-time analytics workloads, which we considered in the logistics and supply chain vertical (see Section~\ref{sec:industry_motivation}) but arguably representative of a wider range of industry data. These workloads call for a contextual data analysis, sometimes entering into the real-time machine learning space, in addition to giving significant attention to out-of-order data point arrival in event streams. They also typically require the reconciliation of numerous event streams, some of which may carry contradictory (erroneous) information.

\begin{figure*}[!!t]
  \centering
  \includegraphics[width=0.7\linewidth]{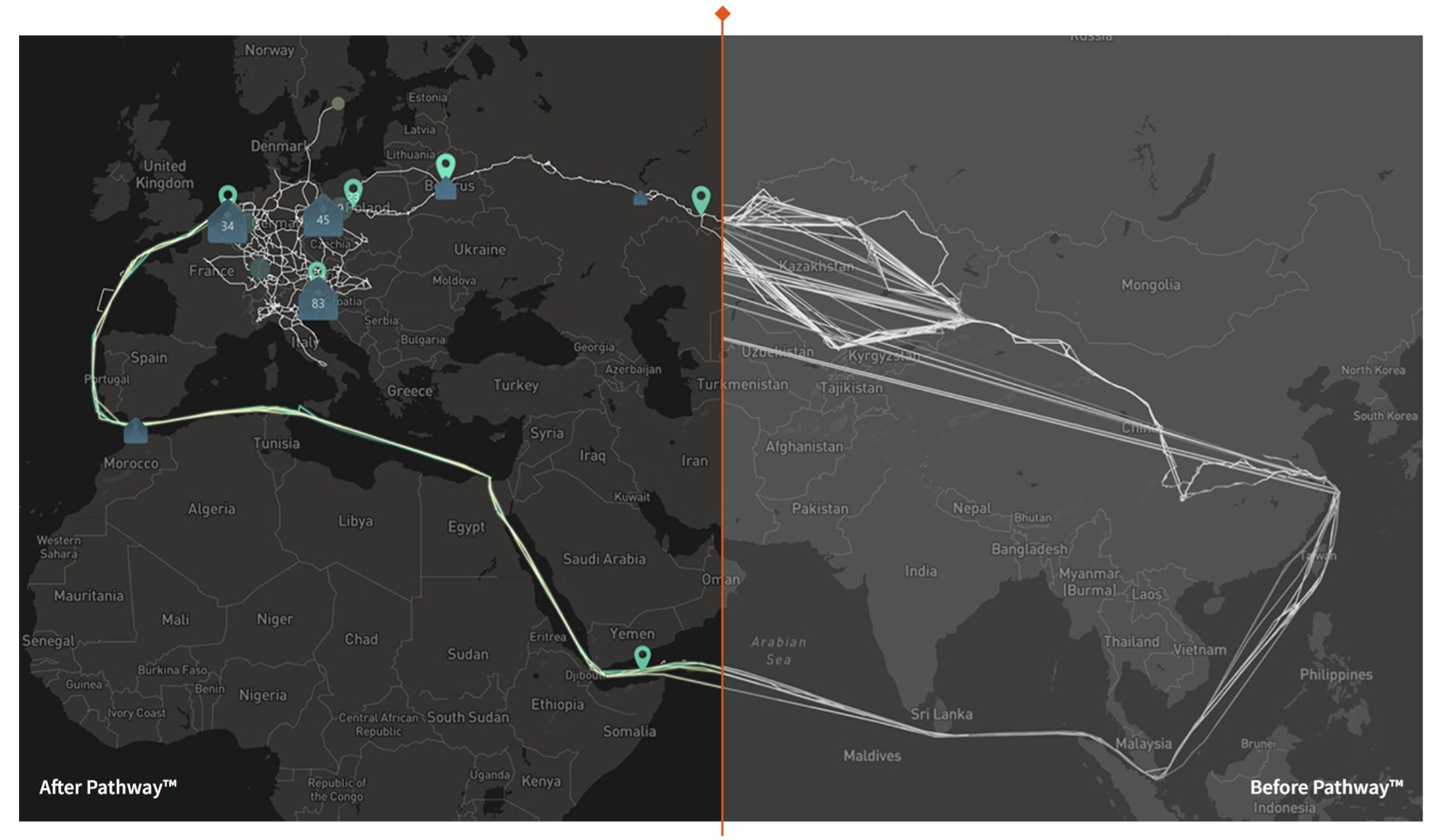}
  \caption{Example of a geospatial event data enrichment process for the Pathway framework. A stream of raw asset locations originating from moving IoT devices [right] is converted into a structured real-time view of processes [left], enabling anomaly-detection and alerting for assets, routes, and locations. The enriched data encompasses a custom map induced from the IoT data that combines typical trajectories (graph edges) and key locations (geofences marked visually either location markers, or clusters of geofences that indicate their count). %
  }
\label{fig:enrichment}
\end{figure*}

Pathway is a data processing framework with a Python API and a reactive data processing engine with a tunable batch size which allows it to be dynamically adjusted for a desired throughput vs latency trade-off. In this paper, we describe its features and provide benchmark results, comparing Pathway to leading batch and streaming data processing systems on a classical analytical benchmark, namely a word counting task implemented using groupby, and a fundamental iterative graph processing algorithm, namely PageRank. We demonstrate that Pathway is capable of achieving throughput outperforming state-of-the-art general purpose batch engines, while being able to respond with latencies better than state-of-the-art streaming systems. At the same time, Pathway succeeds in handling in streaming mode types of iterative and contextual workloads, such as PageRank on a changing graph, which to our knowledge are not supported by any generally industrialized system with a Dataframe/Table API programming layer. Overall, the system performance is owed to a combination of design choices around mapping between tabular syntax and actual key-value data organization, the performance of the underlying differential dataflow assembly for key-value data, the designed operator primitives, and inter-operator optimizations made in the transpilation process.

The paper is structured as follows. In Section~\ref{sec:industry_motivation}, we outline the original motivation for the creation of Pathway originating from industry use cases. In Section~\ref{sec:related}, we provide a historical overview of some of the major approaches to dataflow-based streaming data processing. An overview of the Pathway framework, together with code examples, is put forward in Section~\ref{sec:pathway}. In Section ~\ref{sec:benchmark} we provide results of benchmarks of Pathway, Spark, Flink, and other frameworks for wordcount (streaming) and PageRank (batch, streaming, and backfilling). Section~\ref{sec:conclusion} contains some concluding remarks and directions of work in progress.

\section{Initial industry motivation}\label{sec:industry_motivation}

We initially designed Pathway as a data processor able to accommodate the needs of our real-time analytical platform. We were working with major clients in logistics and supply chain, primarily in international trade (third-party logistics, containerized maritime trade) and postal services. We were creating deployments for perpetual workloads, with a focus on geospatial events data representing events recorded by moving assets. This included IoT data from physical tracker devices, GPS data collected from applications deployed in operators’ mobile phones, location scans recorded at depots and warehouses, and manual operator entries. Physical aspects considered included containers, shipments, trucks, human operators, and vehicles in passenger transportation. A single data enrichment pipeline was deployed and intended for use across multiple use cases in the organization (see Fig.~\ref{fig:enrichment} for an illustration), most immediately in the context of physical process observability and monitoring (real-time control towers), anomaly-detection, and forecasts (Estimated Times of Arrival). This fits into the spirit of building cross-organizational “data products” which are often sponsored by multiple users across the organization \cite{desai_better_2022}.  A discussion of some of the business objectives behind specific deployments undertaken by La Poste (French Postal Services), DB Schenker (a major global freight forwarder), and others can be found on Pathway’s website \cite{pathway}. 

A characteristic shared across most of the deployments we observed, and perhaps a generally-accepted property of the industry verticals concerned, was the nature of the event streams. While the majority of data volumes was arriving with a latency of up to several seconds from measurement to system ingress, a significant part of the data was nonetheless arriving delayed or revised (corrected) within a time window of several hours. This stemmed from numerous causes, from network connectivity issues of devices to manual correction of incorrect or incomplete manual data entries. Overall, the type of data treated gives rise to the following feasibility thesis about process observability design in general: preliminary conclusions (insights and alerts) can be drawn from the incoming data within seconds of data points’ arrival, however, delayed data points can also lead to a change in detected anomalies and changes of “digital twin” representation of the traced processes, even up to 48 hours later.  

In terms of requirements on outcomes, the considered analytics deployments operate across multiple time scales. Certain types of urgent alerts (such as physical assets veering off-course, unplanned door-opening alerts, etc.) need to be raised and dealt with within seconds. Alerts related to transport quality monitoring (temperature changes of perishables, etc.) or changes to congestion in routing and depots typically need to be raised within minutes.  Some information related to changes of process (e.g., detection of new key points of interest) can be recomputed with a significantly large delay, up to 24 hours. This creates room for a certain trade-off between real-time and minibatch computation. 

At the same time, for the analytics data pipeline we were providing to our clients, full batch recomputation of the entire pipeline was not a feasible option. Many insights relied on contextual computation. For example, in the absence of data labels, obtaining insights related to correct transport monitoring conditions (temperature, shocks, etc.) is often only feasible through analysis of “typical” values observed for similar transports in the past. Also, the attribution of certain data anomalies, such as a GPS tracker on an asset leaving a prescribed depot area at an incorrect time, to a specific probable cause (e.g., sensor measurement aberration due to GPS noise, versus, actual transport anomaly) is highly dependent on the context of other measurements made in a similar or comparable area, or by the same or comparable measurement device in the past.

Overall, the data pipelines considered relied on data-driven insights which were fed typically upwards of 1 year of historical data as a reference frame for applied machine learning models. The computational data pipelines despite significant optimization could take upwards of 24 hours to perform a complete recomputation on the multi-core machines allocated to them by the clients. This made our move to streaming / minibatch analytics pipelines driven by a double motivation. First, we needed to reduce latency for the time-critical use cases. Secondly, we needed to introduce pipeline incrementality to optimize excessive recomputation in view of the available computational resources. A final need was related to quickly accommodating changes to pipeline code, notably, being able to integrate fixes to bugs and data problems, without having to wait more than a day to obtain effects. 
For some of the less time-critical and less computationally demanding operations in the pipeline, non-incremental batch computation remained an option. In this case, our expectation was to share logic used by both these batch and real-time procedures, to avoid some of the issues of lambda architecture.
Finally, the engine powering the pipeline was required to be deployable both in client data centers or with major cloud providers, without cloud provider lock-in.
These aspects were the first motivation in our design of the Pathway data processing framework. 

\section{Related frameworks, batch, and streaming systems}\label{sec:related}
MapReduce \cite{dean2008mapreduce} demonstrated how many high-throughput, high-reliability batch computing jobs may be implemented using a common pattern of chaining two user defined operations: a mapper and a reducer, while handling at the framework level the scheduling of computations, their restarts and proactive control over stragglers. It influenced all following data processing systems and defined the main desiderata for systems: decoupling storage from compute and preserving timeliness and reliability of computations in the presence of failing machines.

Flume \cite{chambers2010flumejava} introduced computation graphs on top of MapReduce, allowing deployment of complex multistep data processing pipelines expressed using simple operations on large collections of data. Then, Spark \cite{zaharia2010spark} introduced computation graphs over resilient in-memory datasets, RDDs, for fast and reliable batch computations. Spark pipelines can be written in Java, Python and SQL. Spark was also extended to handle simple cases of streaming data \cite{zaharia2013discretized}, implemented using the RDD model conceptually as an infinite sequence of machines holding batches of a table which come online one-by-one. Independently, Spark GraphX \cite{xin2013graphx} extension generalizes the Pregel \cite{malewicz2010pregel} model for batch computations on graph data.

A major trouble related to stream processing frameworks is state management: beyond the simplest cases, stream elements are not processed independently from one another. Instead, a stream processor maintains a state and updates it with each new consumed stream element. The MillWheel system \cite{akidau2013millwheel} introduced a model for persistent stream processing backing the state to an external scalable database, such as BigTable \cite{chang2008bigtable}  or Spanner \cite{corbett2013spanner}. Concurrently, the Flink \cite{carbone2015apache} framework uses barrier snapshotting to maintain consistent snapshots of the system operation. However, streaming systems must also deal with the problem of producing consistent outputs under data late arrivals. The Dataflow \cite{akidau2015dataflow} model highlighted the problem and presented practical ways to trade correctness for response latency by introducing a notion of watermarks and activation triggers which can be used to precisely control how an application reacts to late data.

The need to provide real-time answers has motivated the Lambda \cite{lambda} architecture. Introduced by Apache Storm \cite{storm} creators, it advocated combining a batch system that processed all historical data in an exact way with a speed layer that provided approximate answers to real-time implemented using an event streaming system. Many systems took the idea further, introducing Streaming and Batch APIs similar enough to be interchangeable based on the needs. The Beam project provides a compiler for computation graphs into either streaming or batch processing backends. Apache Flink maintained batch and streaming engines with similar capabilities, with the notable difference of iterations supported only by the batch layer. Recently, the batch engine was abandoned in favor of running the streaming engine in a batch runtime context which, however, lacks iterative computation support. In this context, we note that Pathway enforces and enables strict parity between batch and streaming computations, by using a unified execution engine.

A powerful incremental data processing framework was developed by the Naiad \cite{murray2013naiad} team and later continued in Rust as a pair of projects, Timely and Differential Dataflow \cite{mcsherryDD}. They propose to generalize tracking progress in a distributed computational graph using partially ordered clocks (Timely Dataflow), and adding to this the capacity of working with deltas (Differential Dataflow). These capacities are also to varying degrees exploited by projects transpiling to Differential Dataflow from different API’s, such as those based on SQL \cite{mcsherry2022materialize} or Datalog \cite{ryzhyk2019differential}. 

Pathway is also inspired by deep learning frameworks: Theano \cite{bergstra2010theano}, Tensorflow \cite{abadi2016tensorflow} and PyTorch \cite{paszke2019pytorch}. Like Pathway, they are libraries for the Python programming language, betting on Python becoming the language for data wrangling and artificial intelligence. The lessons learned from Tensorflow’s story across its versions 1 and 2 were behind a lot of design decisions in Pathway’s dataflow graph-building, as well as Pathway’s decision to adhere to Python as the primary supported API. At the same time, Pathway more closely follows PyTorch’s approach of being able to interact with a partially (dynamically built) computation graph during early experiments. 

\section{The Pathway framework}\label{sec:pathway}

Pathway is a Python-based data processing framework which allows expressing data transformations. Pathway was made to be easy and fast. The implementation has two layers - a runtime engine written in Rust and a Python layer handling computation graph (dataflow) building and optimizing. Pathway code can be written and tested interactively (even in Jupyter notebooks) with the computation graph being built in the background. This allows for quick prototyping needed in data science tasks. The same code can then be run on streaming data sources by disabling eager computation and instead handing the computation graph to the runtime engine.

In any deployment - on either streaming data sources or or in interactive mode, Pathway calls into the same Rust runtime. This allows us to obtain the needed computational efficiency while preserving the fast-paced development of Python code, and also preserving consistency of results across different modes of deployment.

In terms of API, a Table is the primary object for expressing data transformations in Pathway. A Table is a collection of columns that share an underlying set of identifiers. Internally, the Pathway engine operates on columns, which are a data storage with a homogenous type. Columns are indexed with identifiers.

In terms of usability, Tables are directly comparable to Dataframes in Pandas and PySpark, or Tables in Flink’s Table API. From a user perspective, this allows to interpret table transformations as happening on static data, with static columns and tables, while keeping in mind that the expressed computation supports dynamic data. That is, inserts, deletions and modification of input data points are automatically propagated through the dataflow, resulting in updates to outputs. 

Pathway's approach is meant to eliminate most concerns about impact of system processing time and out-of-order data arrival on computational outcomes. Bounded stream workloads can in principle be replayed in Pathway, obtaining repeatable (identical) final results in output tables each time. This comes
subject to a number of evident assumptions - such as not calling in user-defined code external state processing functions or API's which do not behave deterministically, which depend on wall clock time, or on data processing order. Code written natively in Pathway is transparent to these concerns, allowing for predictable operation with clean code logic, and easier testing in CI/CD pipelines. 

Functions defined at data row level may be expressed in Pathway using standard map (apply) and flatmap syntax. A notable extension of Pathway’s syntax is that of transformer classes which take a declarative (ORM) view of computed data tables. They allow for the expression of logic with inter-row dependencies at row-level, permitting e.g. recursive search over rows of a data table representing a graph or other data structure. This general concept is reminiscent of Pregel-like interfaces.

A syntax restriction of Pathway with respect to local batch frameworks such as Pandas is that offset-based row indexing is not supported by Pathway. Instead, Pathway provides fast support for local iteration over sorted indexes (most efficient over columns arriving in almost-sorted order, such as event time columns). This restriction is logical from the point of view of deploying code in a streaming environment, where offsets in data ordering have little meaning, particularly for out-of-order data. It also seems logical for batch workloads which are meant to shard / scale. 

Pathway code transpiles from Python to a high-level dataflow graph which corresponds roughly to the code logic defined by the user. This high-level dataflow then transpiles to an internal dataflow assembly. In a precompilation phase, any SQL expressions which were placed within the Python code via Pathway’s SQL API \cite{pathway} are also first decomposed with a parse tree and converted into the high-level dataflow graph.

In terms of the assembly layer, with respect to incrementally maintaining the results of iterative algorithms, we originally found the logic of the Differential Dataflow project to be closest to satisfying our needs. At the same time, the approach did not provide the data abstraction layer we needed nor interoperability with other systems or data sources. Currently, Pathway code transpiles to a dataflow assembly in Rust relying on a modified subset of Differential Dataflow. A lot of Pathway’s dataflow assembly consists of custom operators, built on top of a modified LSM tree implementation, with extensions allowing for fast handling of event contexts in sharded time-sorted event streams. 

Within the scope of the transformative syntax at a Table level, the correspondence between Pathway’s Table operators and those in its dataflow assembly layer is in general many-to-many, and also affected by optimization settings. The recursion-friendly transformer class syntax at data row level is also transpiled in a special way to the assembly layer. 

In the process of deployed code execution, Pathway stays mostly in its Rust layer. Callbacks through bindings from Rust to the actual Python interpreter occur only when Pathway’s transpiler is unable to eliminate them through one of several applied approaches (i.e., as a fallback of last resort). The scale of Python-related efficiency issues has consequently turned out limited, and so we have so far not needed to deploy Pathway with GIL-free versions of Python.

Pathway is equipped with a connector layer meant for easy use from Python. It provides a mix of input/output interfaces using built-in connectors in the Rust/C layer (for interfacing with Kafka, database CDC over Debezium, file storage formats), as well as configurable Python layer connectors (covering e.g. REST API and websockets). Convenience wrappers around these connectors may be used to perform “asynchronous” callbacks to external API’s. While dataflow-driven, from a usability perspective, the Pathway framework provides a hybrid dataflow/event-driven feel in the connectivity layer, providing a way to describe triggers on events that occur in the data sources, such as new data arrival or timer expiration.

Pathway’s engine predictably relies on uniform workers (each comprising i/o and computational threads), each performing the same workflow, with data sharding. We remark that in enterprise deployments, Pathway’s dataflow is logically extended by a data persistence layer, a control layer based on Kubernetes, and functionalities related to space optimization which are outside the scope of this work; a typical deployment scheme is presented in Fig.~\ref{fig:architecture} in the Appendix.

Changes to Pathway's output Tables are then propagated to down-stream systems through Pathway's output connectors, configured in the same Python code. In the case of commercial deployments of Pathway, output tables may also serve to provide consistent data snapshots to outside systems through an SQL server layer as shown in Fig.~\ref{fig:architecture}.

Unlike most streaming frameworks, Pathway provides streaming consistency guarantees stronger than eventual consistency, and avoids approximate watermarking. For the simple case of a system with a single, non-sharded input data source, with input messages considered atomic (transactional), the user has full control over the stream progress (offsets) for which outputs must be computed by Pathway. This happens by way of COMMIT-type control messages injected through inputs. More generally, Pathway treats in a rigorous way multiple distributed data sources, each with its own timestamped control messages, and harmonizes them into a single output data source which follows an internal Pathway data clock; we omit the details from this paper. 

\subsection{Pathway code examples}

\subsubsection{Simple Wordcount}
The code below demonstrates a complete example for a word counting task:

\begin{lstlisting}[language=Python,breaklines=true]
import pathway as pw
# Kafka settings
rdkafka_settings = {
   "bootstrap.servers": "address:9092",
   ...
}
# Kafka connector fetches json inputs on the "words" topic
words = pw.kafka.read(
   rdkafka_settings,
   topic_names=["words"],
   value_columns=["word"],
   format="json",
   # This setting controls the Pathway batch size
   autocommit_duration_ms=1000
)
# Actual wordcount computation
result = words.groupby(this.word).reduce(
    this.word,
    count=pw.reducers.count(),
)
# Kafka connector writes back computed word counts
pw.kafka.write(result, rdkafka_settings, topic_name="word_counts", format="json")
# Launch the computation
pw.run()
\end{lstlisting}

The code essentially performs four operations. First, it configures how input data is accessed by Pathway. In this case, we use a Kafka connector with JSON message encoding. The autocommit duration determines data batching and controls the tradeoff between system throughput and latency. Next, the actual computation is defined, using a groupby-reduce construct. The third step indicates how Pathway will send its results - again we stream the data to a Kafka topic. The first three steps build a computation graph which encodes all operations that will be performed during execution. The final line of the file starts the computation and enters an infinite loop processing the unbounded data stream.
\subsubsection{PageRank}\label{sec:pagerank}
We demonstrate below a basic implementation of a PageRank computation in Pathway, taken from the documentation~\cite{pathway_pagerank}.

\begin{lstlisting}[language=Python,breaklines=true]
def pagerank(edges, steps: int = 5) -> pw.Table[Result]:
    in_vertices = edges.groupby(id=edges.v).reduce(degree=0)
    out_vertices = edges.groupby(id=edges.u).reduce(degree=pw.reducers.count(None))
    degrees = pw.Table.update_rows(in_vertices, out_vertices)
    base = out_vertices.difference(in_vertices).select(flow=0)

    ranks = degrees.select(rank=6_000)


    for step in range(steps):
        outflow = degrees.select(
            flow=pw.if_else(
                degrees.degree == 0, 0, (ranks.rank * 5) // (degrees.degree * 6)
            ))
        inflows = edges.groupby(id=edges.v).reduce(
            flow=pw.reducers.int_sum(outflow.ix[edges.u].flow)
        )
        inflows = pw.Table.concat(base, inflows)
        ranks = inflows.select(rank=inflows.flow + 1_000)
    return ranks
\end{lstlisting}

The PageRank routine takes as an input a single table of edges, with two columns `u` and `v` containing pointers (hashed vertex labels) to respective endpoints, and returns table indexed by vertices, with the respective number of PageRank surfers computed. 
In the implementation we see operations that transform columns, eg: \texttt{select} which is used for row-wise transformation, \texttt{groupby-reduce} used for aggregation (from edges to vertices), or operations which allow for manipulation of sets of rows in tables (called universes) like \texttt{difference} or \texttt{update\_rows}.
Pathway uses \texttt{ColumnExpressions} as a basic construct for expressing operations on columns. Anatomy of expression is as follows: basic constructs are references to columns, accessed via python attributes of tables, i.e. \texttt{table.name} and consts. Basic arithmetic and logic operators applied on expressions are building blocks for higher order expressions, and are used to describe complex vectorized operations. 

The indexing operator \texttt{table.ix[]} demonstrates Pathway’s capabilities to work with pointers. The .ix operator denotes vectorized dereference, i.e. a join between a key expression and ids of a table. Pointer support facilitates algorithmic thinking during development, internally it is implemented using join, which is of course also fully supported by Pathway.

Pathway uses strong typing of its columns, expressions and tables. Specifically, the type of data stored in each column is tracked. Additionally, row keys (universes) are tracked and used in the type system.

\section{Benchmarks}\label{sec:benchmark}
We report performance of Pathway on streaming and batch tasks which are representative of workloads we want to support: online streaming tasks and graph processing tasks. We evaluate the graph processing task in three modes: first, as a batch computation. Next, as an incremental online computation that should auto-update its results while streaming changes to the graph. Finally, a mixed batch-online mode we call backfilling which evaluates the ability of the engine to switch from batch to online. During backfilling the engine first computes the solution on a large data set in batch mode, then has to start responding in real time to streaming updates. The backfilling scenario simulates e.g. recomputing results after a change to the algorithm.
\subsection{Experiment design}
All code needed to reproduce the benchmarks is publicly available at our GitHub repository\footnote{\url{https://github.com/pathwaycom/}}.

All experiments were run on dedicated machines with: 12-core AMD Ryzen 9 5900X Processor, 128GB of RAM and SSD drives. For all multithreaded benchmarks we explicitly allocate cores to ensure that threads maximally share L3 cache. This is important, as internally the CPU is assembled from two 6-core halves, and thread communication between halves is impacted. For this reason we report results on up to 6 cores for all frameworks.

We run all experiments using Docker, enforcing limits on used CPU cores and RAM consumption.
\subsection{Streaming benchmark: wordcount}

The benchmark task is a simple variant of the Wordcount benchmark in which each line of the input contains a single word, and the goal is to maintain for each word the total number of occurrences in the input stream. In particular, we do not require the code to split the sentences into words, nor to be case insensitive. Instead, we focus on comparing input/output efficiency and the performance of groupby-count operations.

We compare all tested frameworks using the same benchmarking harness which collects all relevant statistics. Test runs are orchestrated using docker compose, which manages all computations: the Kafka service, the data producer and result gathering sink, as well as the wordcount computation to be tested. We manually assign non-overlapping sets of CPU cores to each service. Last, we rely on timestamps assigned by Kafka (using the LogAppendTime option) to compute output latencies.

The streamer is a single-thread Rust binary which simulates a bursty stream with a predefined mean throughput. We have noticed that some frameworks respond with a high latency to the initial messages they produce, but after a while converge to a steady-state performance. To eliminate benchmarking the transient behaviors we employ a burn-in period during which the message throughput is gradually increased. We then discard these events from latency statistics computation. The input Kafka topic is not partitioned.

\subsubsection{Test scenario}

For our usual test scenario, we use 76 million words taken uniformly at random from a dictionary of 5000 random 7-lowercase letter words. We split the dataset into two parts: we use 16 million words as some kind of burn-in period, and we include towards the final readouts only the latencies of the remaining 60 million words. Each experiment is repeated 5 times; the median of all runs is reported.

Since many of the tested frameworks employ minibatching, and guarantee only eventual consistency, we must accordingly define latency. First, we match each entry of the input stream with the earliest value in the output stream that has ‘correct’ count - if there is no exact match, we take the first entry in the output stream that has a larger count. We then compute the latency as the difference of Kafka broker timestamps of matched messages. This way of matching messages produces a matching that is optimistic for systems under consideration. Even if a benchmarked framework processes messages out-of-order, our matching is optimistic and minimizes maximum reported latency. Therefore, we believe the proposed latency definition allows a fair evaluation of all systems using a black-box approach.

\subsubsection{Benchmarked systems and their setup}

In our experiments, we tested several frameworks for stream processing. Each of those frameworks has multiple parameters and variants that allow a user to adjust the performance of the framework to the task at hand. Below, we briefly describe our configuration, and explain why we choose it for each of the frameworks. 

\paragraph{Pathway Setup}
For Pathway we control the minibatch size using the connector autocommit parameter (we have tested 5ms, 10ms, 20ms and 100ms the range is expanded to allow comparison with the two Flink setups) and set the number of threads to match the allocated core count.
 
\paragraph{Flink Setup}
We consider two Flink setups:
Flink defaults, which are implemented using the Scala stream API and which process each message separately,
Flink minibatching in which we configure the runtime to process minibatches of messages (similarly to KafkaStreams, Spark, and Pathway), implemented using Flink Scala Table API.
The two variants provide a control of the throughput-latency tradeoff: the default setup provides better latency at low throughputs, while minibatching obtains maximum sustainable throughput.

We set the \emph{parallelism} parameter to match the number of allocated cores. The minibatching setup controls batch size using the \emph{mini-batch.allow-latency} parameter, whose best settings chosen for comparison were 20ms and 100ms. We have compared Flink performance under the cluster, and single-machine multithreaded configuration and run all tests in the multithreaded configuration which was faster.
\paragraph{Kafka Streams Setup}
We tune two parameters: the size of a batch (using  \emph{autocommit frequency} parameter set to 20ms and 100 ms) that is ingested by Kafka Streams and parallelism. Since Kafka Streams benefits from multithreaded computations only when reading from multi-partitioned topics, we increase its parallelism instead using %
the number of replicas set through docker compose.
Finally, we set the \emph{enableObjectReuse} flag to speed Flink operations by disabling copies of objects made by default for code safety.

\paragraph{Spark Structured Streaming}
Spark supports two streaming modes: Structured Streaming and Continuous Streaming. However, at the time of preparing the benchmarks Continuous Streaming didn’t support groupby aggregations and we have restricted ourselves to benchmark Structured Streaming.

Similarly to other frameworks, we configure parallelism (using \emph{spark-submit --master local[k]}) and batch size (setting \emph{.trigger(Trigger.ProcessingTime(s"\$pTime milliseconds")}) to 20ms or 100ms).
\subsubsection{Wordcount benchmark results}

The Wordcount task is submodular: with large batch sizes reduce the amount of work needed and improve throughput while increasing latency. Thus the task nicely demonstrates the tension between throughput and latency of streaming systems.

\begin{figure*}[h]
  \centering
  \includegraphics[width=\linewidth]{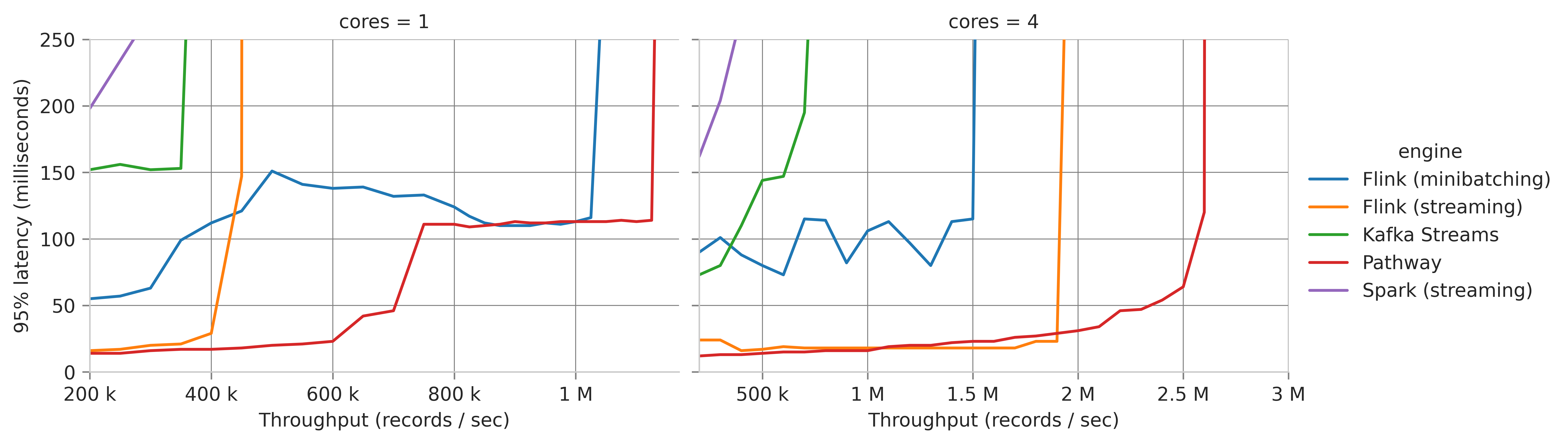}
  \caption{Comparison of latency in the wordcount benchmark for different values of throughput between Flink (default streaming and minibatching setup), Kafka Streams, Spark Structured Streaming, and Pathway. Plots show the 95th percentile of latency and may be considered a representative choice among the percentiles we measured in the benchmark (80th to 99th).}
\label{fig:wordcount}
\end{figure*}

We present experimental results of the observed latency / throughput curve in Fig.~\ref{fig:wordcount}. 
Out of the four tested solutions, Flink and Pathway obtain results on the Pareto front, clearly dominating Spark Structured Streaming and Kafka Streams. Pathway clearly dominates the default Flink setup in terms of sustained throughput, and dominates the Flink minibatching setup in terms of latency for all of the throughput spectrum we could measure. Actually, for most throughputs, Pathway also achieves lower latency than the better of the two Flink setups.

\subsection{Batch benchmark: PageRank}
We benchmark the total time to complete a batch PageRank computation on the LiveJournal social network dataset graph, which contains 4,847,571 nodes and 68,993,773 edges. 

We then compare implementations in Pathway, Flink, and Spark. We rely on an equivalent, idiomatic implementation of PageRank for all three frameworks - noting that In Spark we benchmark several code variants for PageRank: two different flavors of the reference implementation (RDD code, Spark SQL code), and also an implementation proposed in Spark documentation examples, and finally Spark’s GraphX implementation, which are not strictly equivalent.

For all frameworks the input is encoded in JSON format. The output is streamed to /dev/null in order to reduce the influence of non-essential IO operations. The result of the benchmark is the total time spent between the start and the end of the PageRank program. Each experiment is repeated 5 times; the median of all runs is reported.

\paragraph{Pathway setup}
For Pathway we use the implementation provided in its official documentation and described above in section \ref{sec:pagerank}. We report the results of two versions of Pathway engine: 1) the publicly available package, and 2) a build with more aggressive optimizations scheduled for general availability upon extensive testing (the benchmark release).
\paragraph{Flink setup}
For Flink, we implement the same idiomatic logic into Flink’s Scala Table API. The resulting code is slightly more verbose than in Pathway, due to the lack of join-based operators in Flink allowing updating tables more conveniently. Nevertheless, the implementation is analogous to Pathway’s allowing for fair comparison.
\paragraph{Spark setup}
Spark provides GraphX, a dedicated library for graph processing. We included the recommended GraphX PageRank routine in the comparison. In an effort to benchmark similar operations as for the other frameworks, we also implemented the same logic as for Flink and Pathway using the RDD and SQL API's. We have found the SQL API to be about 3 times faster, which we attribute to optimizations enabled by using a more declarative specification of computations; subsequently, we do not present the results for the RDD API. Finally, we also benchmarked a PageRank Implementation provided in the Spark documentation.
\subsubsection{Batch PageRank Results}

We report the batch PageRank results in Table~ \ref{tab:pagerank}. The fastest performance is achieved by the Spark GraphX implementation and the more aggressively-optimized Pathway build. The formulation (and syntax) of the GraphX algorithm is different from the others. Performing an apples-to-apples comparison of performance of equivalent logic in Table APIs, Pathway is the fastest, followed by Flink and Spark.

\begin{table*}
  \caption{ Results of a batch benchmark for PageRank across Flink, Spark, and Pathway. We report the total running time in seconds to process the dataset. The standard code logic is an idiomatic (join-based) implementation. Additionally, two incomparable implementations marked with (*) are benchmarked for Spark.}
  \label{tab:pagerank}
\begin{tabular}{@{}lllrrrr@{}}
\textbf{Framework} & \textbf{Engine runtime}      & \textbf{PageRank code logic} & \textbf{1 core} & \textbf{2 cores} & \textbf{4 cores} & \textbf{6 cores} \\ \midrule
\textbf{Flink}     & batch                        & standard                     & 215                                 & 132                                  & 96                                   & 83                                   \\ \midrule
\textbf{Spark}     & batch                        & standard                     & 556                                 & 340                                  & 193                                  & 143                                  \\
\textbf{}          &                              & from documentation*          & 2643                                & 1408                                 & 781                                  & 571                                  \\
\textbf{}          &                              & graphx routine*               & 130                                 & 78                                   & 44                                   & 35                                   \\ \midrule
\textbf{Pathway}   & unified                      & standard                     & 171                                 & 95                                   & 54                                   & 47                                   \\
\textbf{}          & unified, benchmark release** & standard                     & 108                                 & 63                                   & 44                                   & 39                                   \\
\end{tabular}
\end{table*}

\subsection{Streaming benchmark: PageRank}
We now compare two variants of computing PageRank on a dynamically changing dataset. In the first scenario, called minibatch streaming, we start with an empty graph and add edges in batches of size 1000.

In the second scenario, called backfilling, the system first processes a large batch which contains a significant fraction of the full graph. Then remaining edges are added in batches of size 1000. This scenario tests the ability of the systems to recompute in batch the results as necessary e.g. after a code change, but keeping all state necessary to resume later streaming operation.

The streaming benchmark demands more RAM memory than the batch case - all streaming operations must keep their state in memory for the whole duration of the computation. Thus, we report results on subsets of the  live journal dataset containing 400k and 5M edges. For comparison, we also provide timings of batch runs on these reduced datasets.

\begin{table*}
  \caption{Benchmark results of streaming PageRank computation. We report the total running time in seconds to process the dataset by updating the PageRank results every 1000 edges.}
  \label{tab:online_pagerank}
\begin{tabular}{@{}llrrrr@{}}
\textbf{Dataset}                                                                              & \textbf{Framework} & \textbf{1 core} & \textbf{2 cores} & \textbf{4 cores} & \textbf{6 cores} \\ \midrule
\multirow{2}{*}{\begin{tabular}[c]{@{}l@{}}livejournal, first\\ 400,000 edges\end{tabular}}   & Flink (streaming)  & 91.0                                & 56.0                                 & 29.0                                 & 18.0                                 \\
                                                                                              & Pathway            & 2.2                                 & 1.3                                  & 0.7                                  & 0.6                                  \\ \midrule
\multirow{2}{*}{\begin{tabular}[c]{@{}l@{}}livejournal, first\\ 5,000,000 edges\end{tabular}} & Flink (streaming)  & 3350.0                              & 3190.0                               & 1286.0                               & 830.0                                \\
                                                                                              & Pathway            & 61.0                                & 36.1                                 & 21.4                                 & 17.3                                
\end{tabular}
\end{table*}

\begin{table*}
  \caption{Benchmark results of the backfilling PageRank computation. All systems process first a large batch, then stream additions of remaining edges. We report the total running time in seconds.}
  \label{tab:backfilling_pagerank}
\begin{tabular}{lllrrrr}
\textbf{Dataset}                                                                                       & \textbf{Instance type}                                                                                                                           & \textbf{Framework} & \textbf{1 core} & \textbf{2 cores} & \textbf{4 cores} & \textbf{6 cores} \\ \hline
\multirow{6}{*}{\textbf{\begin{tabular}[c]{@{}l@{}}livejournal, first\\ 5,000,000 edges\end{tabular}}} & \multirow{2}{*}{batch}                                                                                                                           & Flink (batch)      & 27              & 17               & 11               & 10               \\
                                                                                                       &                                                                                                                                                  & Pathway            & 13              & 7                & 4                & 4                \\ \cline{2-7} 
                                                                                                       & \multirow{2}{*}{\begin{tabular}[c]{@{}l@{}}backfilling 4,999,999 edges, then streaming\\ 1 edge\end{tabular}}                                    & Flink (streaming)  & 380             & 200              & 110              & 85               \\
                                                                                                       &                                                                                                                                                  & Pathway            & 13              & 8                & 4                & 4                \\ \cline{2-7} 
                                                                                                       & \multirow{2}{*}{\begin{tabular}[c]{@{}l@{}}backfilling 4,500,000 edges, then streaming\\ 500,000 edges with 1000 edges per update\end{tabular}}  & Flink (streaming)  & 411             & 208              & 113              & 101              \\
                                                                                                       &                                                                                                                                                  & Pathway            & 22              & 13               & 8                & 7                \\ \hline
\multirow{6}{*}{\textbf{\begin{tabular}[c]{@{}l@{}}livejournal, all\\ 68,993,774 edges\end{tabular}}}  & \multirow{2}{*}{batch}                                                                                                                           & Flink (batch)      & 215             & 132              & 96               & 83               \\
                                                                                                       &                                                                                                                                                  & Pathway            & 171             & 95               & 54               & 47               \\ \cline{2-7} 
                                                                                                       & \multirow{2}{*}{\begin{tabular}[c]{@{}l@{}}backfilling 68,993,773 edges, then streaming\\ 1 edge\end{tabular}}                                   & Flink (streaming)  & \multicolumn{4}{c}{Did not finish}                                       \\
                                                                                                       &                                                                                                                                                  & Pathway            & 150             & 90               & 58               & 49               \\ \cline{2-7} 
                                                                                                       & \multirow{2}{*}{\begin{tabular}[c]{@{}l@{}}backfilling 68,493,774 edges, then streaming\\ 500,000 edges with 1000 edges per update\end{tabular}} & Flink (streaming)  & \multicolumn{4}{c}{Did not finish}                                       \\
                                                                                                       &                                                                                                                                                  & Pathway            & 266             & 160              & 100              & 83              
\end{tabular}
\end{table*}

In both variants we focus on the total runtime of tested systems. 

We evaluate only two systems on the streaming PageRank task: Pathway and Flink. We don’t test Kafka Streams because it was suboptimal on the streaming wordcount task. Moreover, no Spark variant supports such a complicated streaming computation: GraphX doesn't support streaming, Spark Structured Streaming doesn't allow chaining multiple groupby’s and reductions, and Spark Continuous Streaming is too limited to support even simple streaming benchmarks.

\paragraph{Pathway Setup}
Pathway runs exactly the same code as in the batch section. Pathway allows external control of batch size by injecting explicit “COMMIT” control messages into the data stream. We use this mechanism. 
\paragraph{Flink Setup}
For Flink the code for the batch and streaming mode is mostly the same, thanks to Flink’s unified Table API. However, some changes were needed as some table operations are only available in batch mode: we have replaced the “minus” operator using leftjoin+filter and we have enforced task specific batching as described in the Appendix~\ref{apx:flink_setup}. In scenarios where Flink was unable to perform as expected, we relaxed the batching requirement and attempted a variety of setups which could possibly lead Flink’s streaming engine to completion of the task, in any way.

\subsubsection{Streaming updates results}
We report the results in the Table~ \ref{tab:online_pagerank}. We see that while both systems are able to run the streaming benchmark, Pathway maintains a large advantage over Flink. It is hard to say whether this advantage is “constant” (with a factor of about 50x) or increases “asymptotically” with dataset size. Indeed, extending the benchmarks to tests on larger datasets than those reported in Table~ \ref{tab:online_pagerank} is problematic as Flink’s performance is degraded by memory issues.

\subsubsection{Backfilling}
This scenario greatly reduces the amount of intermediate results that must be computed.
We once again reduce the dataset size to accommodate memory requirements.

As shown in Table~\ref{tab:backfilling_pagerank}. Pathway again offers superior performance, completing the first of the datasets considered approximately 20x faster than Flink. The first large batch is processed by Pathway in times comparable to the pure batch scenario. This makes the backfilling scenario very practical. In this way, Pathway preserves the ease of defining and updating batch pipelines, while being able to use them in the streaming context as well. For backfilling on the complete LiveJournal dataset, Flink either ran out of memory or failed to complete the task on 6 cores within 2 hours, depending on the setup.

\section{Conclusions}\label{sec:conclusion}

Motivated by applications in context-sensitive real-time analytics for the industry, we have put forward Pathway - a unified engine which can switch mid-way from batch processing to streaming. This enables, e.g., data backfilling at speed. 

As confirmed by the outcomes of all of the benchmarks we have performed so far on standard tasks, Pathway deployments are able to consistently achieve performance better than the compared state-of-the-art frameworks with Table/Dataframe API’s. At the same time, Pathway extends the scope of algorithms on dynamically changing data which may be approached with the convenience of expressing logic through a programming interface focused around manipulation of tables and table rows in Python - potentially unlocking new use cases not previously considered in such streaming frameworks. 

It is interesting to reflect on the scope of algorithms and models which can be described in Pathway and benefit from efficient recomputation following data changes. As discussed herein, iterative algorithms on changing data (such as PageRank for a changing graph) fall into this category. A different case concerns event-streams, where changes to the streamed data replay only a relatively recent part of the head of the stream. Convenient handling of such cases of partially frozen data at a syntactic level is foreseen in future revisions of Pathway.

We also foresee next steps around tighter integration of Pathway with Python libraries and API's, to demonstrate how to fully leverage the power of the Python ecosystem in Pathway. Concrete examples include easy-to-setup connectors for external systems with push-API's (especially in monitoring and alerting use cases), and interfacing with Machine Learning libraries suited for working with online data, including time series.

\begin{figure*}[t]
  \centering
  \includegraphics[width=0.7\linewidth]{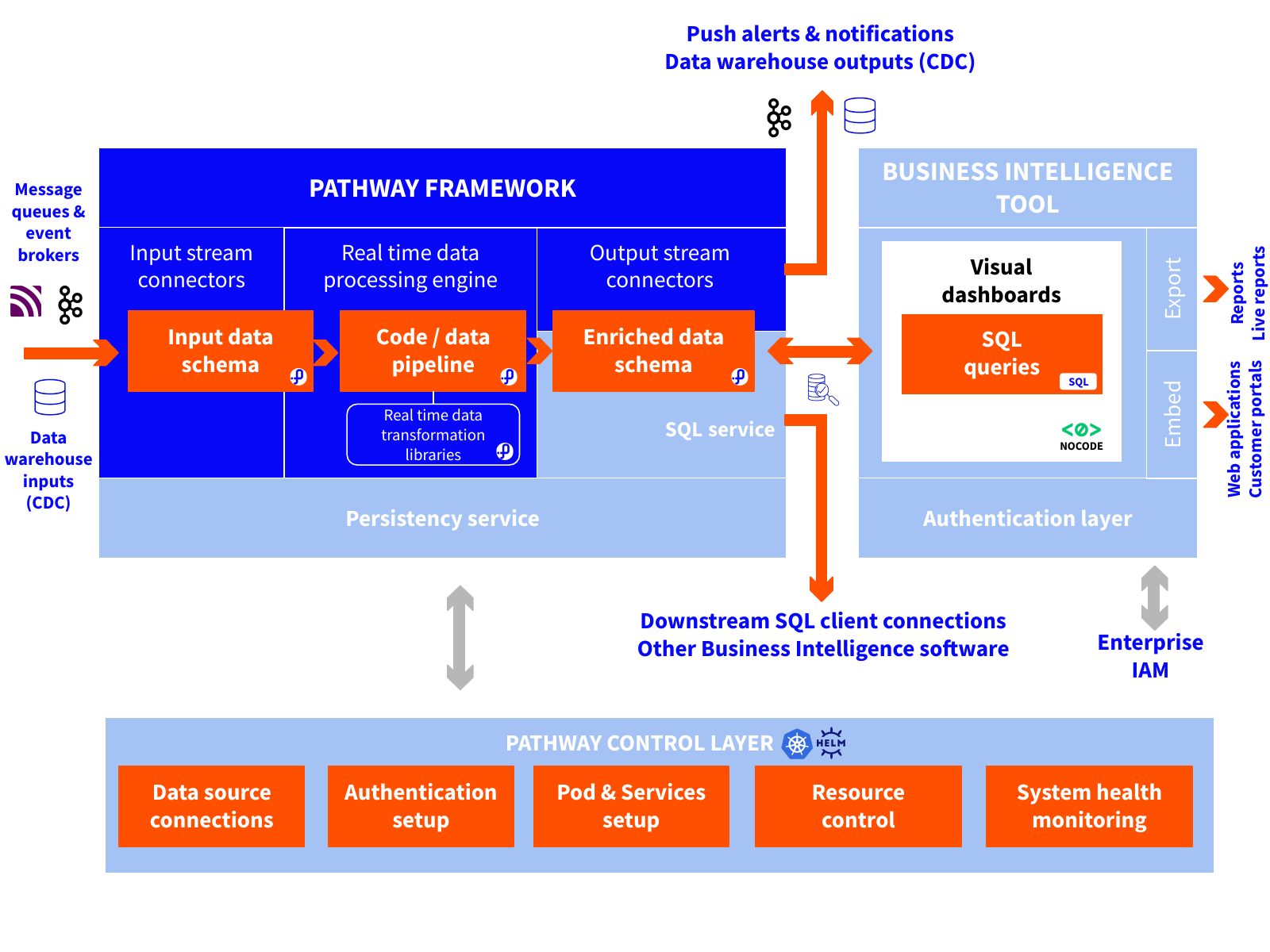}
  \caption{Schema of typical Pathway framework integration in enterprise deployments.}
\label{fig:architecture}
\end{figure*}

\enlargethispage{4mm}
\bibliographystyle{ACM-Reference-Format}
\bibliography{pathway}

\appendix

\section{Consuming mini-batches of a given size in streaming mode in Flink}\label{apx:flink_setup}

In order to achieve commits every 1000 edges we used Flink’s minibatch optimization. To this end we had to configure \emph{table.exec.mini-batch.allow-latency} and \emph{table.exec.mini-batch.size parameters} (both are required by Flink).\balance

Initially, we tried to set \emph{mini-batch.size=1000}, and set a large allow-latency. This however, resulted in updates occurring much more frequently than every 1000 edges. It seems that the mini-batch.size configuration applies to every aggregation operator (and not input), rendering it useless for the PageRank algorithm which performs some aggregations in each iteration.

Therefore, we started tuning the \emph{mini-batch.allow-latency} setting. Flink can work with processing time or event time. It turns out that this processing mode affects the mini-batching mode (ProcTimeMiniBatchAssignerOperator vs RowTimeMiniBatchAssginerOperator, see Flink’s source code). We therefore aimed to set event time for the $k$-th edge to be equal to $k$ with a watermark strategy for ascending timestamps. We set “mini-batch.allow-latency” to 1000 and “mini-batch.size” to something large in order to make it irrelevant -- we used $10^8$.

This however did not work on its own, as Flink’s planner was automatically choosing mode to use processing time, as our PageRank application has only unbounded aggregations and global joins (i.e. we don’t have any windowing in the logic). To circumvent this issue, we added a dummy operator at the beginning of the pipeline which makes an interval join on input edges with self. This forced Flink's planner to use event time (referred to also as RowTime) in minibatching.

\end{document}